\documentclass[sigconf]{acmart}

\usepackage{amsmath}
\usepackage{mathtools}
\usepackage{amsthm}

\usepackage{natbib}
\usepackage{algorithm}
\usepackage{algorithmic}
\usepackage{booktabs}
\usepackage{color}
\usepackage{array}
\usepackage{booktabs}
\usepackage{float}

\usepackage{subcaption}
\AtBeginDocument{%
  }

\setcopyright{acmlicensed}
\copyrightyear{2025}
\acmYear{2025}
\acmDOI{}
\acmConference[MLoG-GenAI '25]{Workshop on Machine Learning on Graphs in the Era of Generative Artificial Intelligence}{August 6, 2025}{Toronto, ON, Canada}

\acmISBN{}

\begin{document}

\title{Spectral Manifold Harmonization for Graph Imbalanced Regression}

\author{Brenda Nogueira}
\affiliation{
  \institution{University of Notre Dame}
  \city{Notre Dame}
  \state{Indiana}
  \country{USA}
}
\email{bcruznog@nd.edu}

\author{Gabe Gomes}
\affiliation{
  \institution{Carnegie Mellon University}
  \city{Pittsburgh}
  \state{Pennsylvania}
  \country{USA}
}

\author{Meng Jiang}
\affiliation{
  \institution{University of Notre Dame}
  \city{Notre Dame}
  \state{Indiana}
  \country{USA}
}
\author{Nitesh V. Chawla}
\affiliation{
  \institution{University of Notre Dame}
  \city{Notre Dame}
  \state{Indiana}
  \country{USA}
}

\author{Nuno Moniz}
\affiliation{
  \institution{University of Notre Dame}
  \city{Notre Dame}
  \state{Indiana}
  \country{USA}
}



\begin{abstract}
Graph-structured data is ubiquitous in scientific domains, where models often face imbalanced learning settings. In imbalanced regression, domain preferences focus on specific target value ranges that represent the most scientifically valuable cases; however, we observe a significant lack of research regarding this challenge. In this paper, we present Spectral Manifold Harmonization (SMH), a novel approach to address imbalanced regression challenges on graph-structured data by generating synthetic graph samples that preserve topological properties while focusing on the most relevant target distribution regions. Conventional methods fail in this context because they either ignore graph topology in case generation or do not target specific domain ranges, resulting in models biased toward average target values. Experimental results demonstrate the potential of SMH on chemistry and drug discovery benchmark datasets, showing consistent improvements in predictive performance for target domain ranges.
Code is available at~\href{https://github.com/brendacnogueira/smh-graph-imbalance.git}{\textit{https://github.com/brendacnogueira/smh-graph-imbalance.git}}.
\end{abstract}




\maketitle

\section{Introduction}
Graph-structured data has become increasingly important in scientific domains, particularly drug discovery, materials science, and genomics. 
Graph Neural Networks (GNNs) have revolutionized the modeling of such data by operating directly on graph structures, enabling more accurate predictions of molecular properties, material characteristics, and biological interactions, for example. In drug discovery alone, GNNs have demonstrated significant promise for tasks such as property prediction \citep{xiong2020attentive}, molecular design \citep{jin2018junction}, and drug-target interaction prediction \citep{lim2019predicting}. The pharmaceutical industry has embraced these methods to accelerate the traditional drug development pipeline, which typically costs over \$1 billion and spans more than a decade from discovery to market \citep{vamathevan2019applications}. 

While considerable research has targeted imbalanced classification problems in graph learning \citep{almeida2024overcoming, xia2024novel}, the regression setting has received little attention \citep{ribeiro2020imbalanced,liu2023semisupervisedgir}. Crucial scientific problems involve predicting continuous properties where the most valuable cases are rare, e.g., in drug discovery, high-potency compounds represent a tiny fraction of the chemical space but are the most scientifically interesting \citep{silva2022model}. Standard machine learning approaches, including GNNs, typically optimize for average performance across the entire distribution, resulting in models that perform poorly on these traditionally infrequent but valuable cases. Additionally, existing oversampling techniques for imbalanced data often fail to preserve the complex topological properties inherent in graph-structured scientific data, thereby limiting their effectiveness in these domains.

\begin{figure}[!h]
\begin{center}
\centerline{\includegraphics[width=0.95\columnwidth]{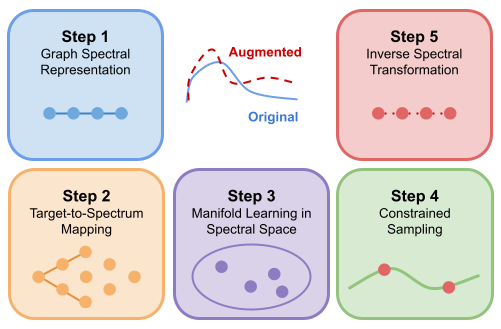}}
\caption{Visual description of the Spectral Manifold Harmonization method's workflow.} 
\Description{}
\label{fig:smh}
\end{center}
\end{figure}

In this paper, we present Spectral Manifold Harmonization (SMH), a novel approach for tackling imbalanced regression~\cite{ribeiro2020imbalanced,Moniz2017ImbalancedRegression,Ma2024ChemistryImbalancedRegression} on graph-structured data. SMH (Figure~\ref{fig:smh}) operates in the graph spectral domain—the eigenspace of the graph Laplacian—to generate synthetic graph samples that preserve essential topological properties while focusing on underrepresented target distribution regions. Building on the concept of relevance in imbalanced regression~\citep{ribeiro2020imbalanced}, which maps target values to non-uniform domain preferences, SMH learns a continuous manifold of valid graph structures by relating target values and the spectral domain, allowing the generation of new samples with targeted properties. 

\paragraph{Novelty.} This approach overcomes the limitations of existing oversampling techniques in regression settings by operating in a space that captures the structural properties of graphs, enabling the generation of realistic synthetic examples that address the imbalance problem without distorting the underlying graph topology.

\paragraph{Findings.} Experimental results show that SMH considerably improves predictive performance on target ranges in benchmark datasets from drug discovery. Specifically, models trained with SMH-augmented datasets improve the accurate prediction of rare compounds, while maintaining or improving performance on average cases. We also demonstrate how synthetic graphs generated by SMH preserve essential structural characteristics of the original data, confirming the effectiveness of our spectral approach.



\section{Related Work}

The challenge of imbalanced distributions in graph learning tasks has received increasing attention, particularly in scientific domains where rare values are critical. Recent research by \citet{almeida2024overcoming} demonstrates that imbalanced learning in drug discovery datasets can be tackled with techniques such as oversampling and loss function manipulation when using Graph Neural Networks (GNNs). Despite these advances, most approaches operate directly in graph space rather than the spectral domain, limiting their ability to maintain global structural constraints. \citet{bo2023survey} published a comprehensive survey on spectral GNNs, highlighting their unique ability to capture global information and provide better expressiveness than spatial approaches. \citet{wang2022powerful} further analyzed the theoretical expressive power of spectral GNNs, proving that they can produce arbitrary graph signals under specific conditions. However, these methods focus on balanced and classification datasets, illustrating the novelty and significance of SMH.

\subsection{Spectral Graph Methods}

Spectral graph theory has a rich history in machine learning, with applications spanning dimensionality reduction, clustering, and graph signal processing. Recent work in spectral methods includes Specformer~\citep{bo2023specformer}, combining spectral GNNs with transformer architectures to create learnable set-to-set spectral filters, or the work by \citep{li2025largescale} to enhance the scalability of spectral GNNs without decoupling the network architecture, addressing a key limitation in previous approaches. These advanced spectral methods demonstrate improved performance on various graph learning tasks, but do not specifically target the regression setting or leverage the spectral domain for manifold harmonization in imbalanced scenarios.
Our SMH method extends these ideas to regression, enabling targeted generation in underrepresented regions while maintaining global graph properties.

\subsection{Manifold Learning for Structured Data}

Manifold learning principles underpin many approaches to generating synthetic structured data. 
Recently, \citet{zhong2024knowledge} described how models can be enhanced by incorporating structured knowledge representations and latent manifold embeddings, in the context of knowledge-augmented graph machine learning for drug discovery. Similarly, \citet{baumgartner2023manifold} demonstrated that incorporating manifold information improves synthetic oversampling techniques for high-dimensional spectral data where standard approaches often fail. Our SMH approach differs from these works by explicitly modeling the regression target-to-spectrum mapping and performing manifold learning in the spectral domain, making it particularly suited for scientific applications with imbalanced regression targets.

\subsection{Graph Sampling and Synthesis in Scientific Domains}

Due to domain-specific constraints and validity requirements, scientific applications pose unique challenges for graph-based methods. \citet{yao2024knowledge} provided a comprehensive bibliometric analysis of GNN applications in drug discovery, showing significant growth in this area and highlighting the need for methods to handle the inherent data imbalances in these domains. Similarly, \citet{fan2024reducing} addressed the challenge of overconfident errors in molecular property classification, demonstrating the importance of uncertainty quantification in imbalanced datasets. These approaches focus primarily on classification rather than regression tasks, and do not specifically utilize spectral representations to address imbalance. 

On regression tasks, a review on GNNs for predicting synergistic drug combinations~\citep{zhang2023review} noted that graph-based models often suffer from imbalanced data distributions, affecting their performance. They emphasized the need for methods to handle such imbalances to improve predictive accuracy effectively. Our SMH method offers a domain-agnostic approach that incorporates scientific validity constraints while focusing on generating underrepresented regions of the target distribution, bridging critical gaps in existing methodologies for imbalanced regression on graph-structured data in scientific applications.

\section{Methods: Spectral Manifold Harmonization}\label{method}

Our Spectral Manifold Harmonization (SMH) method addresses imbalanced regression on graph-structured data by learning to generate synthetic graph samples in underrepresented regions of the target distribution while preserving their topological properties. The key insight is that operating in the graph spectral domain allows us to construct a continuous manifold of valid graph structures, making it possible to sample new graphs with targeted properties. SMH integrates the concept of relevance from recent work on imbalanced regression \citep{ribeiro2020imbalanced, silva2022model} and consists of five main components (Figure~\ref{fig:smh}): we first transform graphs into their spectral representation, learn how target values map to this spectral space with emphasis on relevant regions, model the manifold of valid spectral representations, strategically sample from underrepresented areas, and finally transform back to generate new graph instances that address the imbalance problem.

\subsection{Graph Spectral Representation and Relevance Concept}\label{methods:begin}

Let $G = (V, E)$ be a graph with $|V| = n$ nodes and a set of edges $E$. We define the adjacency matrix $\mathbf{A} \in \mathbb{R}^{n \times n}$ where $A_{ij} = 1$ if $(i,j) \in E$, 0 otherwise, the degree matrix $\mathbf{D}$ with $D_{ii} = \sum_{j} A_{ij}$, and the normalized Laplacian $\mathbf{L}_{\text{norm}} = \mathbf{I} - \mathbf{D}^{-1/2}\mathbf{A}\mathbf{D}^{-1/2}$. The spectral decomposition of $\mathbf{L}_{\text{norm}}$ yields $\mathbf{L}_{\text{norm}} = \mathbf{U}\mathbf{\Lambda}\mathbf{U}^T$, where $\mathbf{U} = [u_1, u_2, ..., u_n]$ contains the eigenvectors and $\mathbf{\Lambda} = \text{diag}(\lambda_1, \lambda_2, ..., \lambda_n)$ contains the eigenvalues with $0 = \lambda_1 \leq \lambda_2 \leq ... \leq \lambda_n \leq 2$. For any graph signal $\mathbf{x} \in \mathbb{R}^n$, its Graph Fourier Transform (GFT) is given by $\hat{\mathbf{x}} = \mathbf{U}^T\mathbf{x}$, where $\hat{\mathbf{x}}$ represents the signal in the spectral domain.

A key concept in addressing imbalanced regression is relevance, which maps target values to non-uniform domain preferences~\citep{ribeiro2020imbalanced}. 
In this context, a continuous, domain-dependent relevance function $\phi(Y): \mathcal{Y} \rightarrow [0,1]$ expresses the application-specific bias concerning the target variable $\mathcal{Y}$. A domain expert ideally defines the relevance function for the specific task where the expert inputs information on the available target value-relevance pairs, i.e., which value is considered low or high-relevance. When this information is unavailable, the function can be interpolated from boxplot-based statistics where extreme values are considered high-relevance and the distribution median is considered the lowest point of relevance.


\subsection{Relevance-Guided Target-to-Spectrum Mapping}

Given a dataset $\mathcal{D} = \{(G_i, y_i)\}_{i=1}^N$ of graph-label pairs, we learn a parameterized function $f_\theta: \mathbb{R} \rightarrow \mathbb{R}^k$ that maps regression target values to spectral coefficients, where $k < n$ is the number of significant eigenmodes. The mapping function is implemented as a neural network:

\begin{equation}
f_\theta(y) = \mathbf{W}_L \cdot \sigma(\mathbf{W}_{L-1} \cdot \sigma( \cdots \sigma(\mathbf{W}_1 \cdot y + \mathbf{b}_1) \cdots ) + \mathbf{b}_{L-1}) + \mathbf{b}_L
\end{equation}

where $\sigma$ is a non-linear activation function, and $\mathbf{W}_l, \mathbf{b}_l$ are learnable parameters. We incorporate the relevance concept into our optimization objective by weighting the loss according to the importance of each target value:

\begin{equation}
\mathcal{L}(\theta) =\frac{1}{N} \sum_{i=1}^N \phi(y_i) \cdot \|\mathbf{s}_i - f_\theta(y_i)\|^2 + \alpha \cdot \Omega(\theta)
\end{equation}

where $\mathbf{s}_i = \mathbf{U}_i^T \mathbf{x}_i$ are spectral coefficients of graph $G_i$, $\alpha$ is a regularization parameter, and $\Omega(\theta)$ is a regularization term. This relevance-weighted loss function ensures that the model focuses more on learning the mapping for high-relevance target values.

\subsection{Manifold Learning in Spectral Space}

We model the distribution of spectral coefficients conditioned on target values as a multivariate Gaussian:

\begin{equation}
p(\mathbf{s}|y) = \mathcal{N}(\mu(y), \Sigma(y))
\end{equation}

where $\mu(y) = f_\theta(y)$ and $\Sigma(y)$ is estimated using a relevance-weighted covariance:

\begin{equation}
\Sigma(y) = \sum_{i=1}^N w_i(y) \cdot (\mathbf{s}_i - \mu(y))(\mathbf{s}_i - \mu(y))^T
\end{equation}

with weights determined by target similarity:


\begin{equation}
w_i(y) = \frac{K(y, y_i)}{\sum_{j=1}^N K(y, y_j)}
\end{equation}

where $K(y, y_i) = \exp(-\gamma(y - y_i)^2)$ is a Gaussian kernel. 
This weighting scheme ensures that the manifold captures the variability in of each region more accurately when modeling the covariance structure.

\subsection{Constrained Sampling for Underrepresented Regions}

To address target distribution imbalance, we first estimate the density $p(y)$ using kernel density estimation:

\begin{equation}
p(y) = \frac{1}{Nh}\sum_{i=1}^N K\Big(\frac{y - y_i}{h}\Big)
\end{equation}

where $K$ is a kernel function and $h$ is the bandwidth parameter. We define a sampling weight function that combines both the inverse density and the relevance:

\begin{equation}
w(y) = \phi(y) \cdot (p(y) + \epsilon)^{-1}
\end{equation}

where $\epsilon$ is a small constant to prevent division by zero. This function prioritizes regions that are underrepresented (low density) and highly relevant. To generate new samples, we:
\begin{enumerate}
    \item Sample target values $y_{\text{new}}$ with probability proportional to $w(y)$
    \item Generate spectral coefficients $\mathbf{s}_{\text{new}} \sim \mathcal{N}(\mu(y_{\text{new}}), \Sigma(y_{\text{new}}))$
\end{enumerate}

%
%
%
%
\subsection{Inverse Spectral Transformation}

Finally, to reconstruct graphs, given $\mathbf{s}$, we:
\begin{enumerate}
    \item Reconstruct spectral representation $\hat{\mathbf{x}} = [\mathbf{s}_{\text{valid}}, \mathbf{0}]$
    \item Apply inverse GFT: $\tilde{\mathbf{x}} = \mathbf{U}\hat{\mathbf{x}}$
    \item Construct adjacency matrix: $\tilde{A}_{ij} = \sigma(\tilde{\mathbf{x}}_i \cdot \tilde{\mathbf{x}}_j)$, where $\sigma$ is a sigmoid function
\end{enumerate}

The resulting graph $\tilde{G}$ preserves essential topological properties while targeting underrepresented yet relevant regions of the target distribution, effectively augmenting the training set to improve regression performance on rare but valuable cases. By integrating the concept of relevance throughout our method, we ensure that the synthetic samples generated by SMH focus specifically on the regions of the target space that are most important for the application domain.

\section{Experiments}

In this section, we evaluate the effectiveness of SMH in generating synthetic samples that preserve key structural patterns from the original molecular dataset and improve prediction performance in domain-relevant target value ranges. 

We address the following research questions:

\begin{enumerate}
    \item Do synthetic graphs generated by SMH follow the molecular structure patterns of the original dataset?
    \item Does the use of SHM improve predictive accuracy in target ranges considered scientifically important?
    \item 
    How do SMH's that focus on specific domain regions impact the overall performance?
    \item How does manifold learning and constrained sampling perform in comparison with traditional augmentation?
    \item How does SMH perform in comparison with pre-trained models?
\end{enumerate}

These questions guide our analysis of the structural fidelity of generated samples and the practical impact of SMH on regression performance across diverse benchmarks.

\subsection{Methods} 
We converted the SMILES into a networkx~\cite{hagberg2020networkx} graph to build the spectral manifold harmonization space. Then, we used XGBoost to train a model to predict the eigenvalues from a given target. For property prediction, we then convert the networkx graph format for \textit{PyTorch Geometric} data format and input it in a Graph Isomorphism Network (GIN)~\cite{xu2018powerful}, a powerful tool for graph-based machine learning tasks due to of its capability to effectively differentiate between different graph structures, using MSE as a loss function. The hyperparameter is presented in Appendix~\ref{app:models} with a 5-fold cross-validation. We also compared our relevance-guided target-to-spectrum transformation and constrained sampling approach with SMOGN~\cite{branco2017smogn}. For the Spectral+SMOGN baseline, we first compute the spectral representation as described in Section~\ref{methods:begin}, and then apply the SMOGN method. The inverse transformation used for decoding remains the same. To compare with a pre-trained model, we used HiMol~\cite{zang2023hierarchical}, which is a framework to learn molecule graph representation for property prediction.

\subsection{Data}
Our experimental evaluation focuses on molecular data, using regression tasks from MoleculeNet~\cite{wu2018moleculenet}: ESOL, FreeSolv, and Lipophilicity (Lipo). The datasets are briefly described in Table~\ref{tab:dataset}. The datasets exhibit a long-tailed distribution toward the lower end of the property range, and we define our relevance function to assign higher importance to these.
\begin{table}[ht]
\centering
\caption{Summary of Molecular Property Datasets}
\label{tab:dataset}
\resizebox{\columnwidth}{!}{
\begin{tabular}{lcc}
\toprule
\textbf{Dataset} & \textbf{\# of Compounds} & \textbf{Description} \\
\midrule
ESOL           & 1{,}128    & Water solubility dataset \\
FreeSolv       & 642        & Hydration free energy of small molecules in water \\
Lipophilicity  & 4{,}200    & Octanol/water distribution coefficient of molecules \\
\bottomrule
\end{tabular}
}
\end{table}

\section{Results and Discussion}
This section addresses the research questions raised in Section~\ref{method}, specifically concerning SMH's ability to generate synthetic graphs and model performance when using the SMH method for generating and leveraging such data.

\subsection{Synthetic Generated Graphs}

\begin{figure}[h]
\begin{center}
\centerline{\includegraphics[width=\columnwidth]{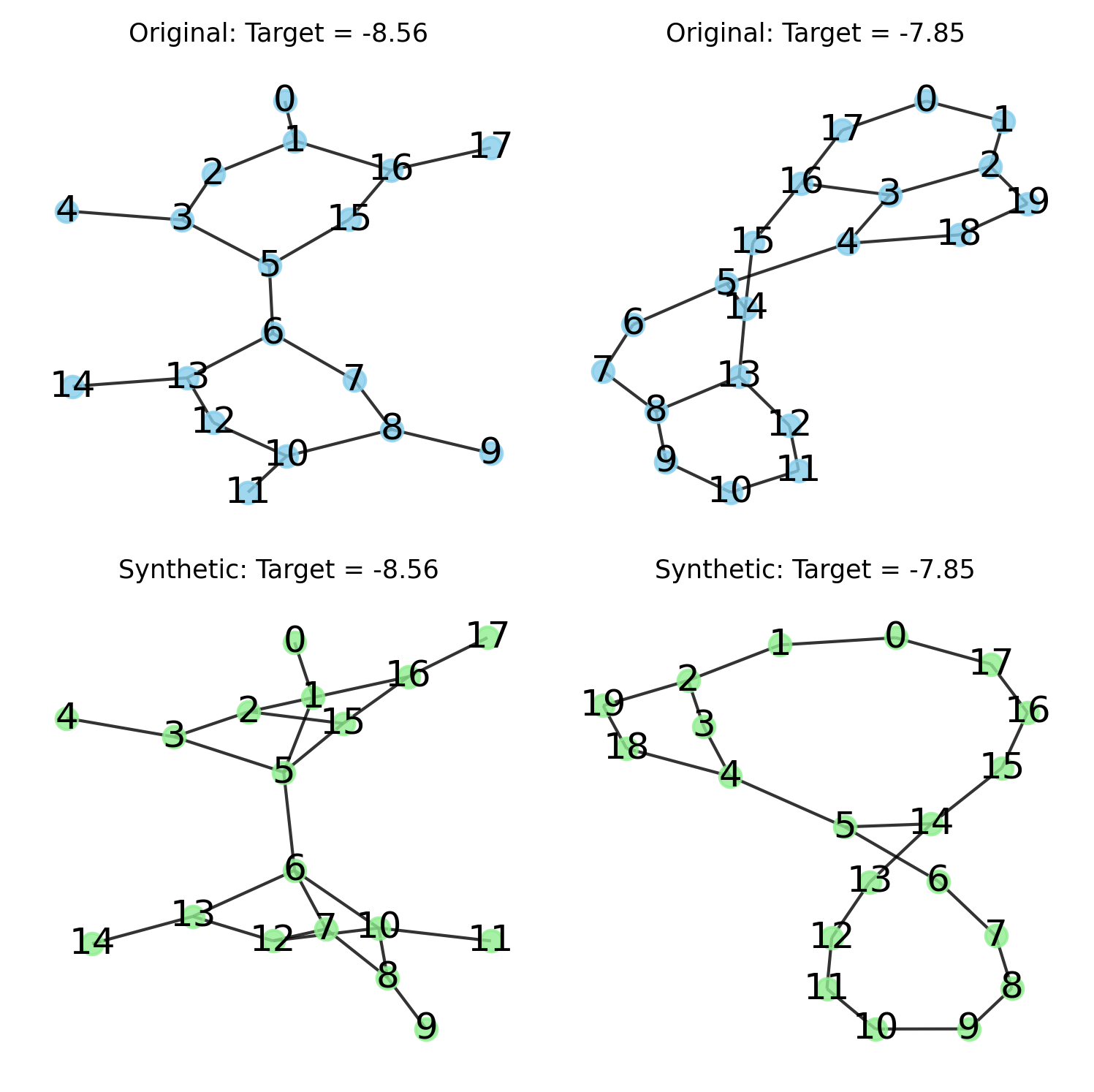}}
\caption{Illustration of graphs selected for augmentation and the corresponding synthetic graphs generated using the approach described in Section~\ref{method}, for the ESOL dataset.}
\label{fig:generated_graphs}
\end{center}

\end{figure}

An illustration of the graphs selected for augmentation and the corresponding synthetic graphs generated using the approach described in Section~\ref{method} is presented in Figure~\ref{fig:generated_graphs}. Results show that the generated samples follow the molecular structure patterns of the original dataset. Importantly, they are not simple copies but exhibit structural variations, indicating that the method produces diverse, meaningful graphs.

\begin{figure*}[htp]
\centering
\begin{subfigure}{1.0\textwidth}
    \includegraphics[width=1.0\linewidth]{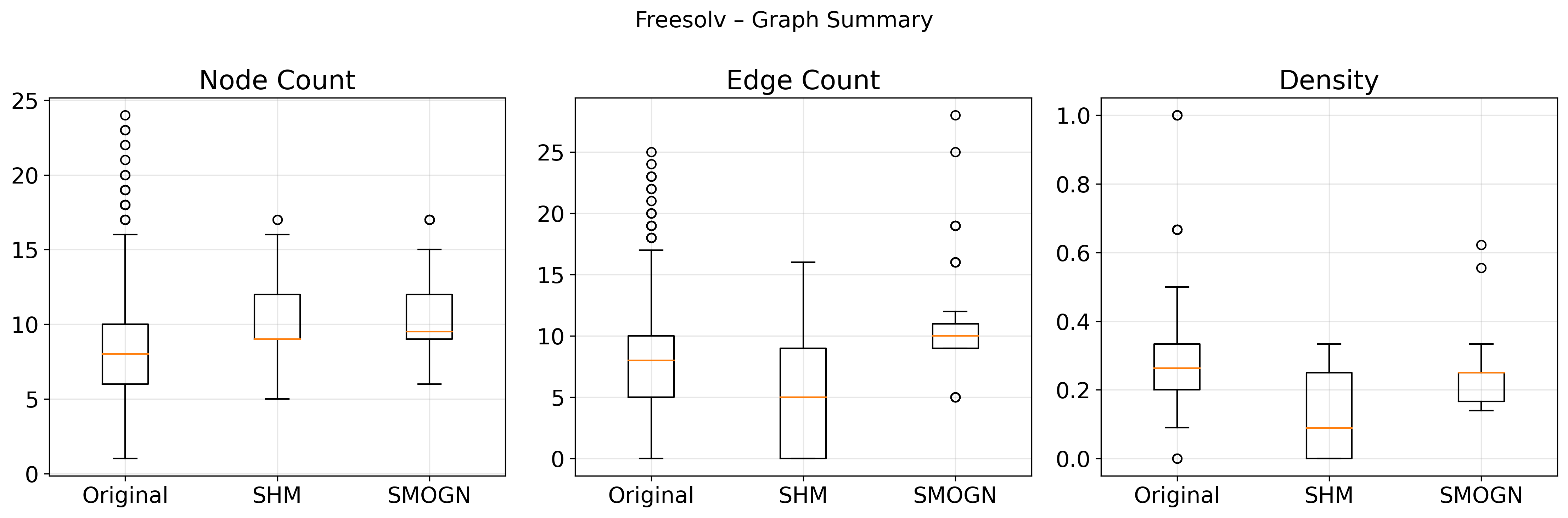}
\end{subfigure}
\begin{subfigure}{1.0\textwidth}
    \includegraphics[width=1.0\linewidth]{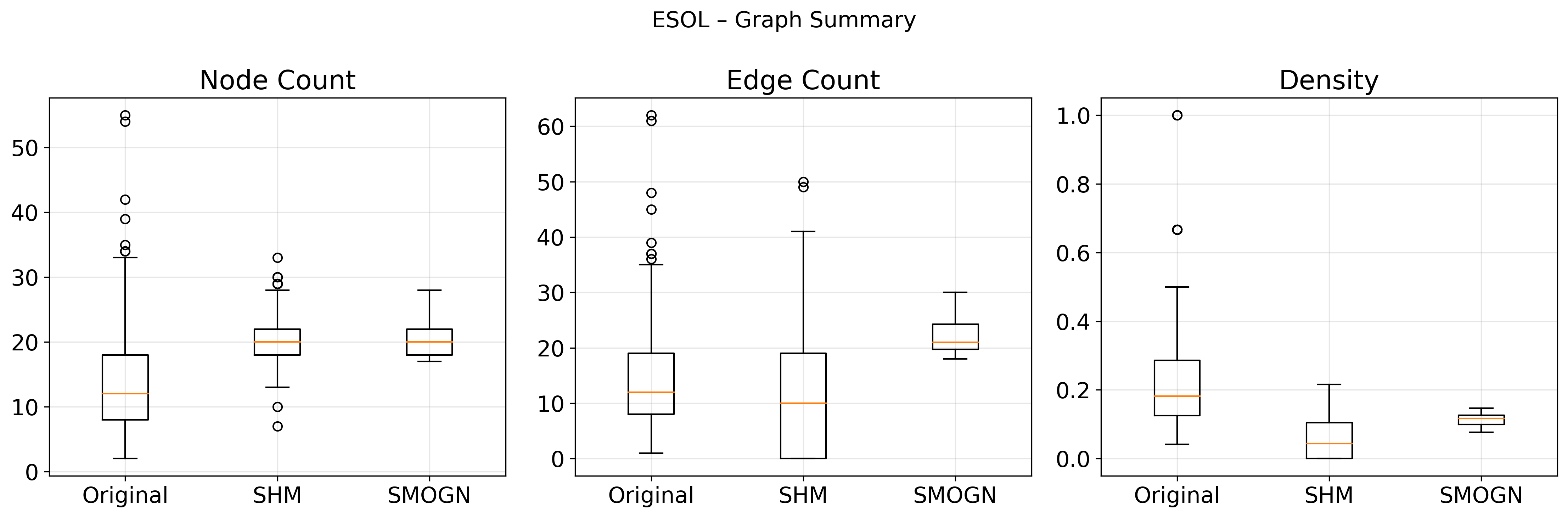}
\end{subfigure}
\begin{subfigure}{1.0\textwidth}
    \includegraphics[width=1.0\linewidth]{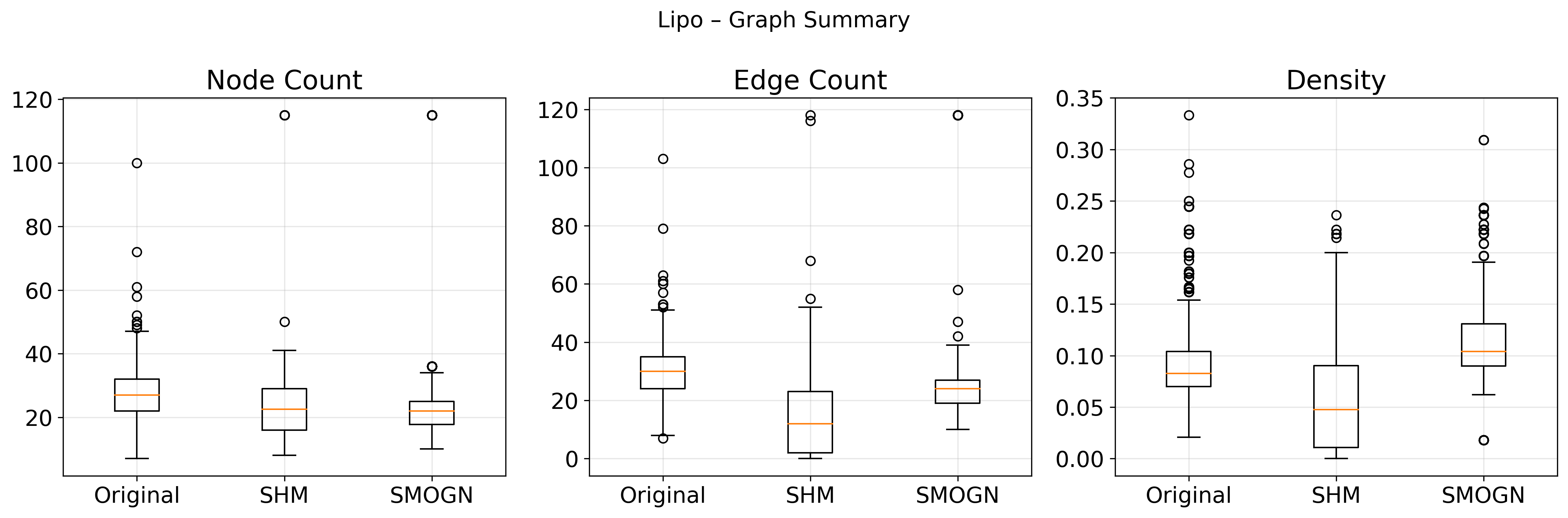}
\end{subfigure}
\caption{Comparison of Mean and Standard Deviation of Node and Edge Counts Between Original and Synthetic Graphs.}
\label{fig:generated_analysis}
\end{figure*}

A comparison of the mean and standard deviation of node and edge counts between the original and synthetic graphs is provided in Figure~\ref{fig:generated_analysis}. The number of nodes remains very similar across sets. Minor differences are observed in the number of edges and graph density, with the synthetic graphs showing slightly higher mean values. However, these differences remain within an acceptable range, supporting the validity of the generated graphs (\textbf{RQ1}). Further validation on the generated graphs can be addressed. 

\subsection{Model Performance}

\begin{figure*}[htp]
\centering
\begin{subfigure}{0.33\textwidth}
    \includegraphics[width=1.0\linewidth]{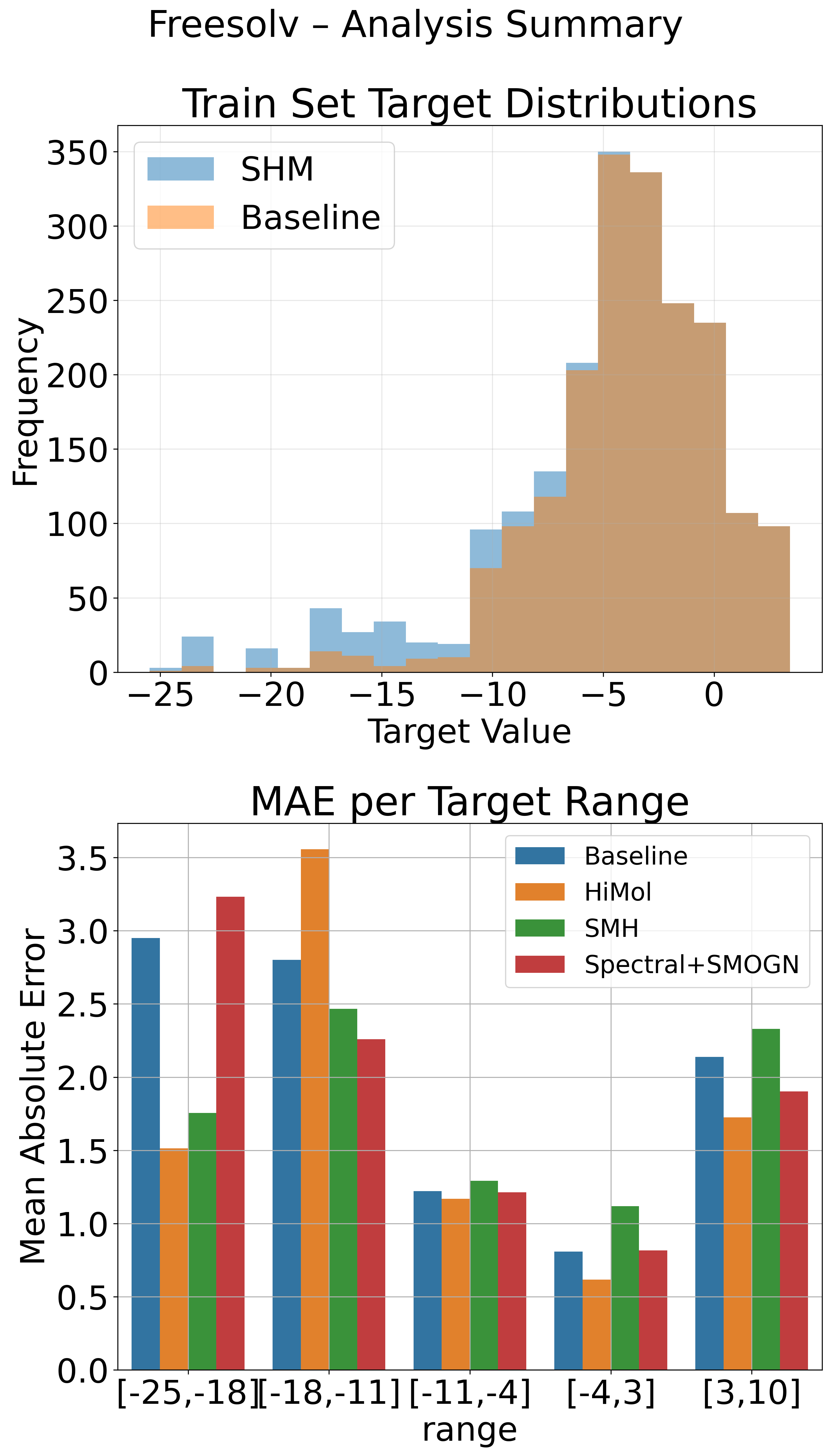}
\end{subfigure}
\begin{subfigure}{0.33\textwidth}
    \includegraphics[width=1.0\linewidth]{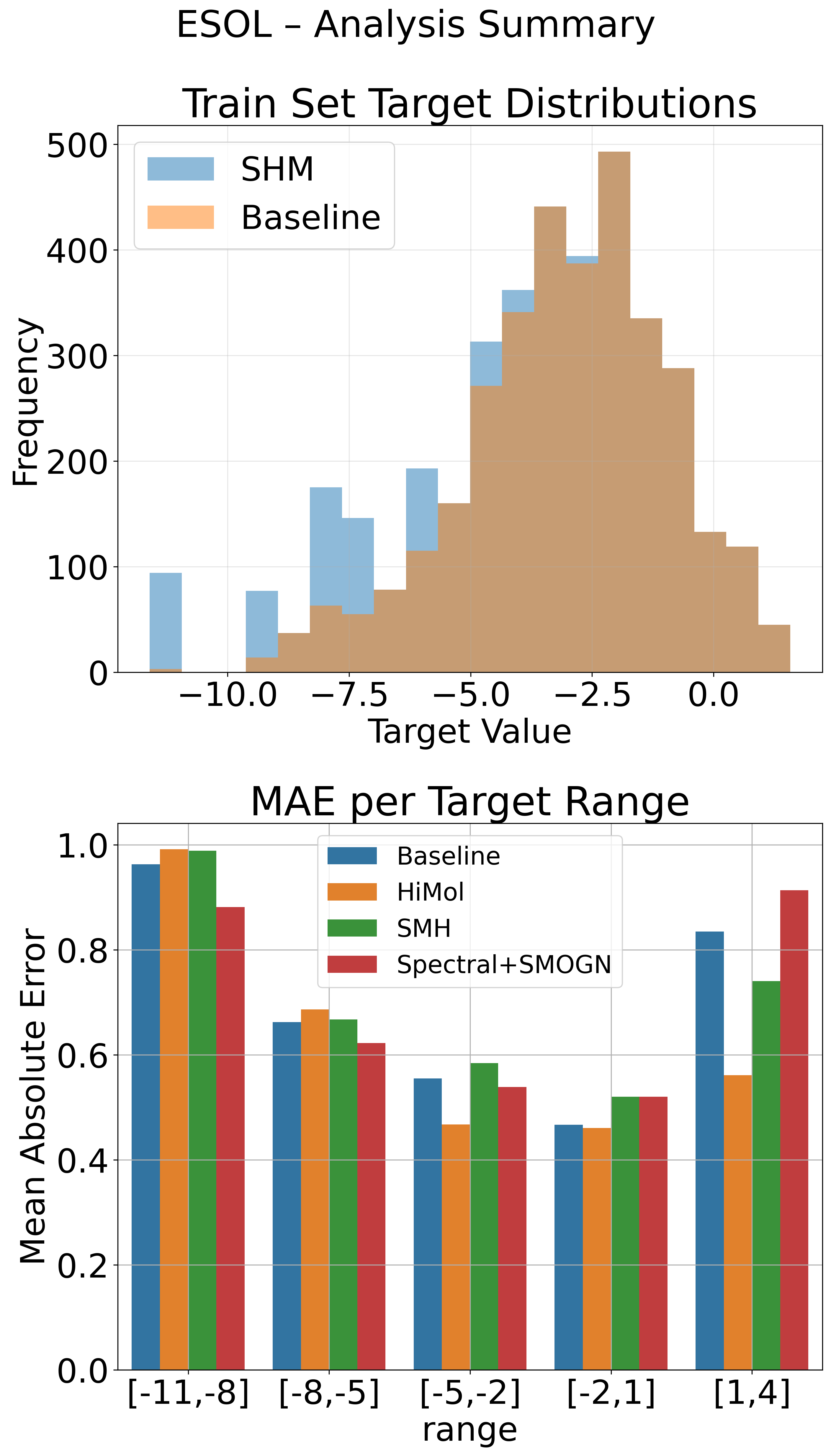}
\end{subfigure}
\begin{subfigure}{0.33\textwidth}
    \includegraphics[width=1.0\linewidth]{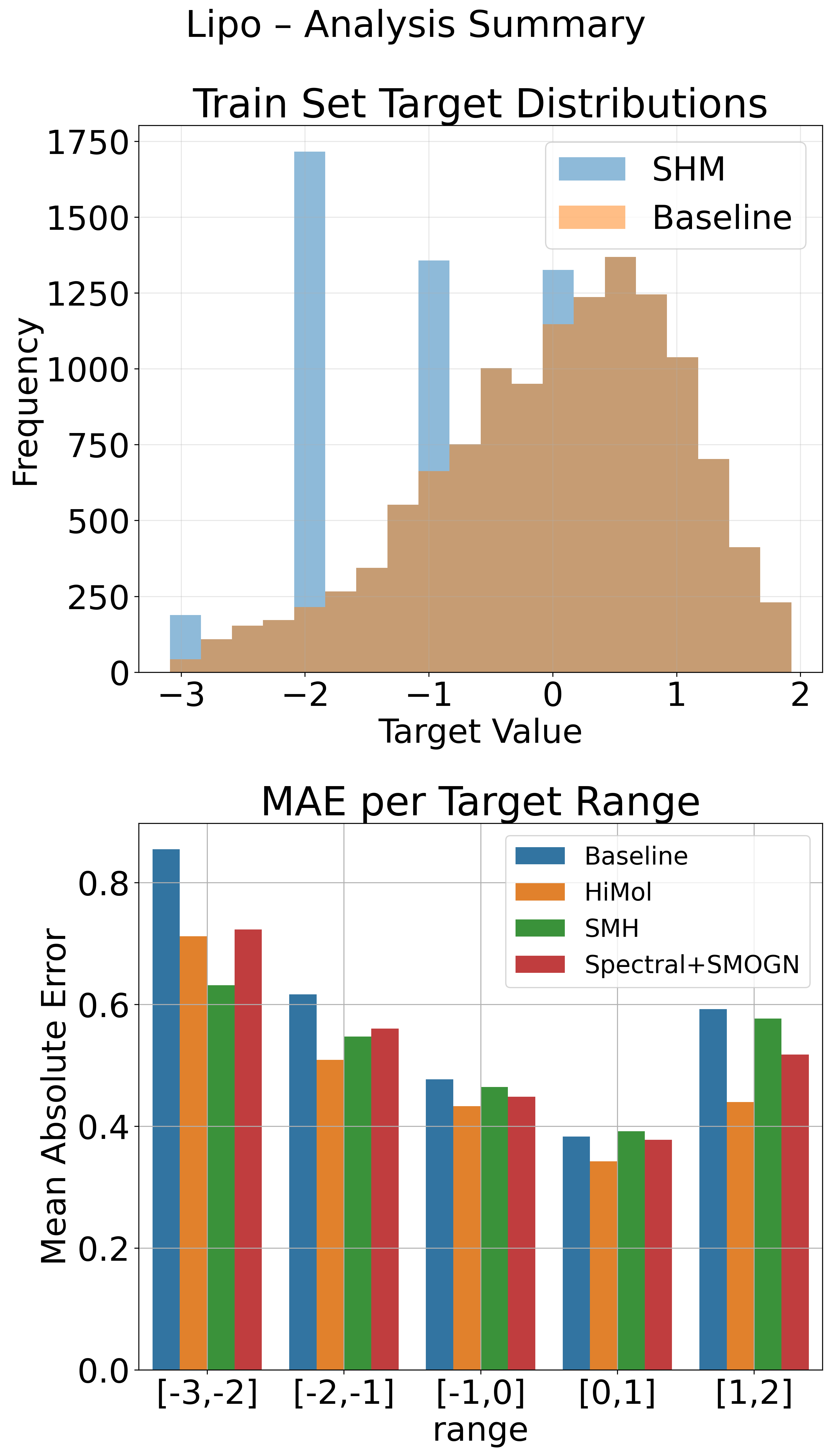}
\end{subfigure}
\caption{Distribution of train dataset with and without synthetic augmentation, along with the improvements for each part of the test set domain, for each dataset.}
\label{fig:dist}
\end{figure*}

The experimental results are reported in Table~\ref{tab:improvements}, and Figure~\ref{fig:dist} illustrates each dataset's improvement across different domain regions. The results show noticeable improvements in the lower range of the domain (\textbf{RQ2}), where our augmentation is focused and where training data is scarce, with minimal or no degradation in the higher range (\textbf{RQ3}). This results in an improvement in the SERA evaluation metric and similar results in other metrics. When compared to Spectral+SMOGN, our method improves performance on the most relevant ranges, demonstrating the effectiveness of manifold learning and constrained sampling in generating augmented graphs and their potential for further improvement (\textbf{RQ4}). In comparison with a pre-trained model, our approach demonstrates very comparable results with significant improvements in the low range part of the domain (\textbf{RQ5}). 

\begin{table}[ht]
\caption{Experimental results for the FreeSolv, ESOL, and LIPO datasets, using the SERA, MAE, RMSE, and $R^2$ evaluation metrics. Arrows signal the direction for best results, also noted in bold.}
\label{tab:improvements}
\begin{center}
\resizebox{\columnwidth}{!}{
\begin{tabular}{lcccc}

\toprule
                &                   &    FreeSolv               &                         \\ \hline
\textbf{Metric} & \textbf{Baseline} & \textbf{SHM} & \textbf{Spectral+SMOGN} & \textbf{HiMol}  \\
\midrule
SERA $\downarrow$ & $0.83\pm0.9$ & $\mathbf{0.55\pm0.35}$ & $0.69\pm0.58$ & $0.71\pm0.93$ \\
MAE $\downarrow$ & $1.07\pm0.16$ & $1.25\pm0.17$ & $1.06\pm0.14$ & $\mathbf{0.95\pm0.17}$ \\
RMSE  $\downarrow$ & $1.67\pm0.33$ &  $1.81\pm0.3$ & $1.59\pm0.32$ & $\mathbf{1.46\pm0.41}$ \\
$R^2$ $\uparrow$ & $0.81\pm0.07$ & $0.77\pm0.11$ & $0.83\pm0.06$ & $\mathbf{0.85\pm0.08}$\\
\toprule
                &                   &    ESOL               &                         \\ \hline
\textbf{Metric} & \textbf{Baseline} & \textbf{SHM} & \textbf{Spectral+SMOGN} & \textbf{HiMol} \\
\midrule
SERA $\downarrow$ & $0.07\pm0.03$ & $0.08\pm0.03$ & $\mathbf{0.06\pm0.02}$ & $0.08\pm0.01$ \\
MAE $\downarrow$  &  $0.56\pm0.05$ & $0.59\pm0.04$ & $0.56\pm0.02$ & $\mathbf{0.51\pm0.02}$ \\
RMSE $\downarrow$  & $0.73\pm0.07$ & $0.77\pm0.05$ & $0.73\pm0.04$  & $\mathbf{0.7\pm0.02}$ \\
$R^2$ $\uparrow$ & $0.87\pm0.03$ & $0.86\pm0.02$ & $0.88\pm0.02$ & $\mathbf{0.89\pm0.01}$ \\

\toprule
                &                   &    Lipo               &                         \\ \hline
\textbf{Metric} & \textbf{Baseline} & \textbf{SHM} & \textbf{Spectral+SMOGN} & \textbf{HiMol}\\
\midrule
SERA $\downarrow$ & $0.11\pm0.03$ & $\mathbf{0.08\pm0.01}$ & $0.09\pm0.02$ & $\mathbf{0.08\pm0.01}$ \\
MAE $\downarrow$  & $0.49\pm0.01$ & $0.47\pm0.02$ & $0.46\pm0.01$ & $\mathbf{0.42\pm0.02}$ \\
RMSE  $\downarrow$ & $0.66\pm0.01$ & $0.64\pm0.02$ & $0.62\pm0.03$ &  $\mathbf{0.57\pm0.01}$ \\
$R^2$ $\uparrow$ & $0.57\pm0.02$ &  $0.6\pm0.03$ & $0.62\pm0.04$ & $\mathbf{0.67\pm0.01}$ \\

\bottomrule
\end{tabular}
}
\end{center}
\end{table}



\section{Conclusion}

In this work, we introduced Spectral Manifold Harmonization (SMH), a novel method for addressing the challenge of imbalanced regression on graph-structured data. By generating synthetic samples in the spectral domain of graphs, SMH maintains topological integrity while focusing learning on underrepresented but domain-relevant target value regions. Our approach bridges a critical gap in the literature by combining domain-specific relevance modeling with structure-preserving augmentation, enabling improved predictive performance in settings such as drug discovery where rare cases are of great interest.

Experimental results on benchmark datasets demonstrate that models trained with SMH-augmented data outperform conventional approaches, particularly in low-frequency target regions, without sacrificing performance elsewhere. Structural analyses confirm that generated graphs remain faithful to the original distribution regarding key topological properties. SMH thus offers a principled and effective augmentation strategy for improving learning in scientific domains where data imbalance and structural complexity often limit model effectiveness. 

\subsection{Future Improvements}
Spectral Manifold Harmonization (SMH) has shown strong potential for addressing imbalanced regression on graph-structured data, but several avenues remain for further enhancement. First, integrating domain-specific constraints into the graph generation process could improve the realism and scientific validity of the synthetic graphs. Second, the absence of semantic context in the current synthesis process limits the interpretability and relevance of the generated data, highlighting the need for a hybrid spectral-semantic approach. Future work will also involve evaluating SMH across a wider range of benchmark datasets and predictive models to further optimize performance. Additionally, we plan to conduct more comprehensive comparisons with existing state-of-the-art methods and expand the application domains beyond drug discovery, including areas such as biology and materials science, to better assess the generalization of our method. To this end, we aim to develop a hybrid framework that integrates semantic information into the generation and modeling pipeline to further enhance prediction performance and scientific relevance.


\bibliographystyle{ACM-Reference-Format}
\bibliography{sample-base}

\appendix

\section{Model Details.}\label{app:models}

This section provides additional information about the models used in our experiments. The hyperparameters tested for XGBoost, the property prediction model, and the augmentation strategies are summarized in Table~\ref{tab:hyperparams}.  Given the relatively small dimensionality of the eigenvalue vectors (fewer than 50 features), XGBoost outperformed neural networks in our evaluations. Model selection was based on performance on the validation split, using the SERA metric.

\begin{table}[h!]
\centering
\caption{Hyperparameter search space}
\label{tab:hyperparams}
\begin{tabular}{cll}
\hline
\textbf{Parameter} & \textbf{Values Tested} \\
\hline

XGBoost & Number of estimators & 10, 50, 100, 250 \\
& Learning rate & 0.001, 0.01, 0.1 \\
& Max depth & 3, 5, 10 \\
\hline

GIN Model & Learning rate & 0.01, 0.005, 0.001 \\
& Batch size & 16 \\
& Hidden dimension & 32, 64 \\
& Number of layers & 2, 5 \\
& Epochs & 500 \\

\hline

SMH & $\gamma$ & 1.0, 0.5 \\
& Augmentation sampling & 0.20, 0.15, 0.10 \\
& Binarization cut-off & 0.3, 0.2, 0.1 \\ \hline
SMOGN & Relevance threshold & 0.95, 0.99 \\
\hline
\end{tabular}
\end{table}

We defined a relevance function $ \phi(y) $ using the \texttt{extremes} method with three control points:

\begin{align*}
\phi(y) = 
\begin{cases}
1       & \text{if } y = \min(\mathcal{Y}) \\
0.025   & \text{if } y = \mu = \mathrm{mean}(\mathcal{Y}) \\
0       & \text{if } y = \max(\mathcal{Y})
\end{cases}
\end{align*}

where $\mathcal{Y}$ denotes the set of target values in the training data. The relevance function smoothly interpolates between these points to emphasize extreme values.

\begin{figure}[H]
\centering
\begin{subfigure}{0.9\columnwidth}
    \includegraphics[width=1.0\linewidth]{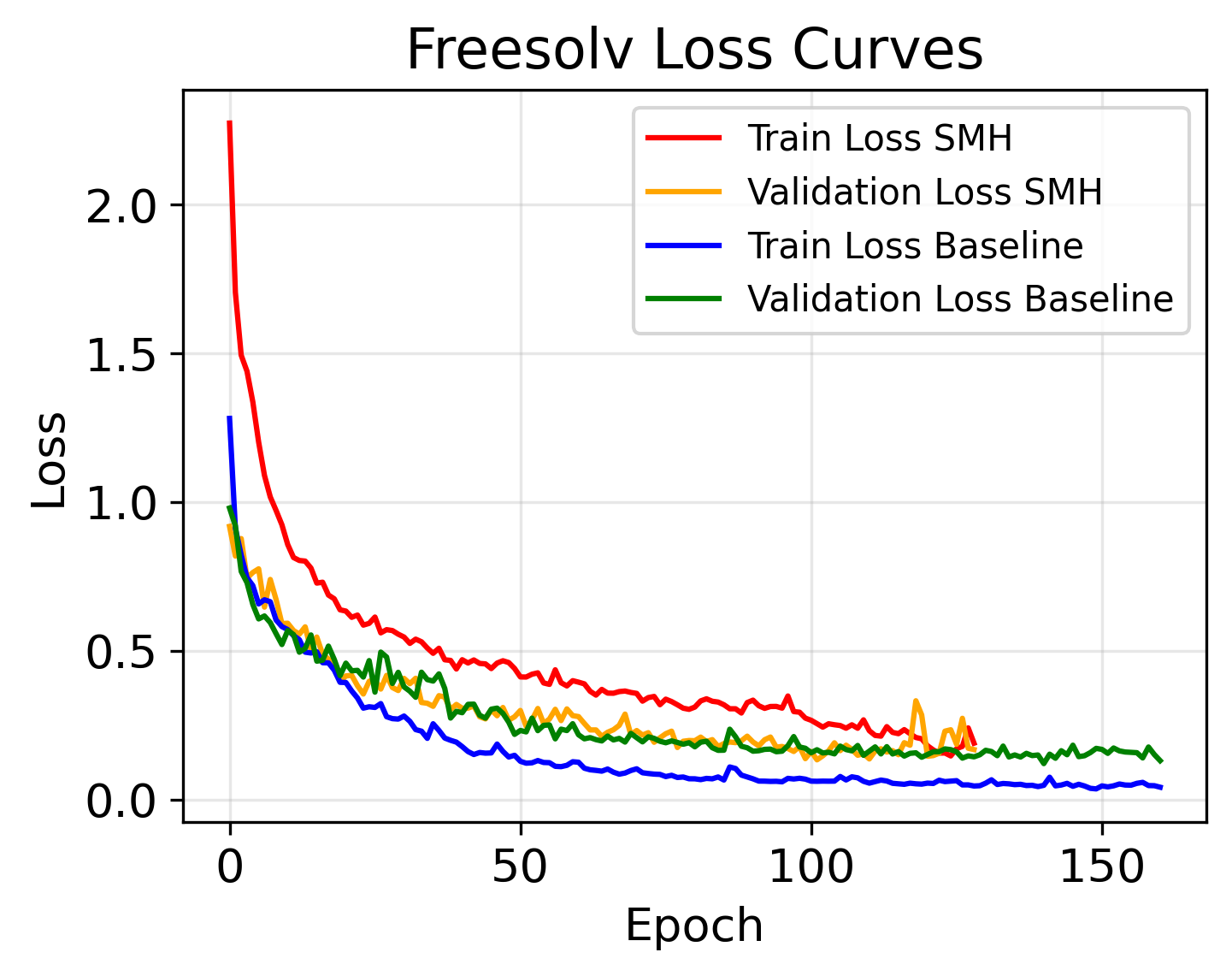}
\end{subfigure}
\begin{subfigure}{0.9\columnwidth}
    \includegraphics[width=1.0\linewidth]{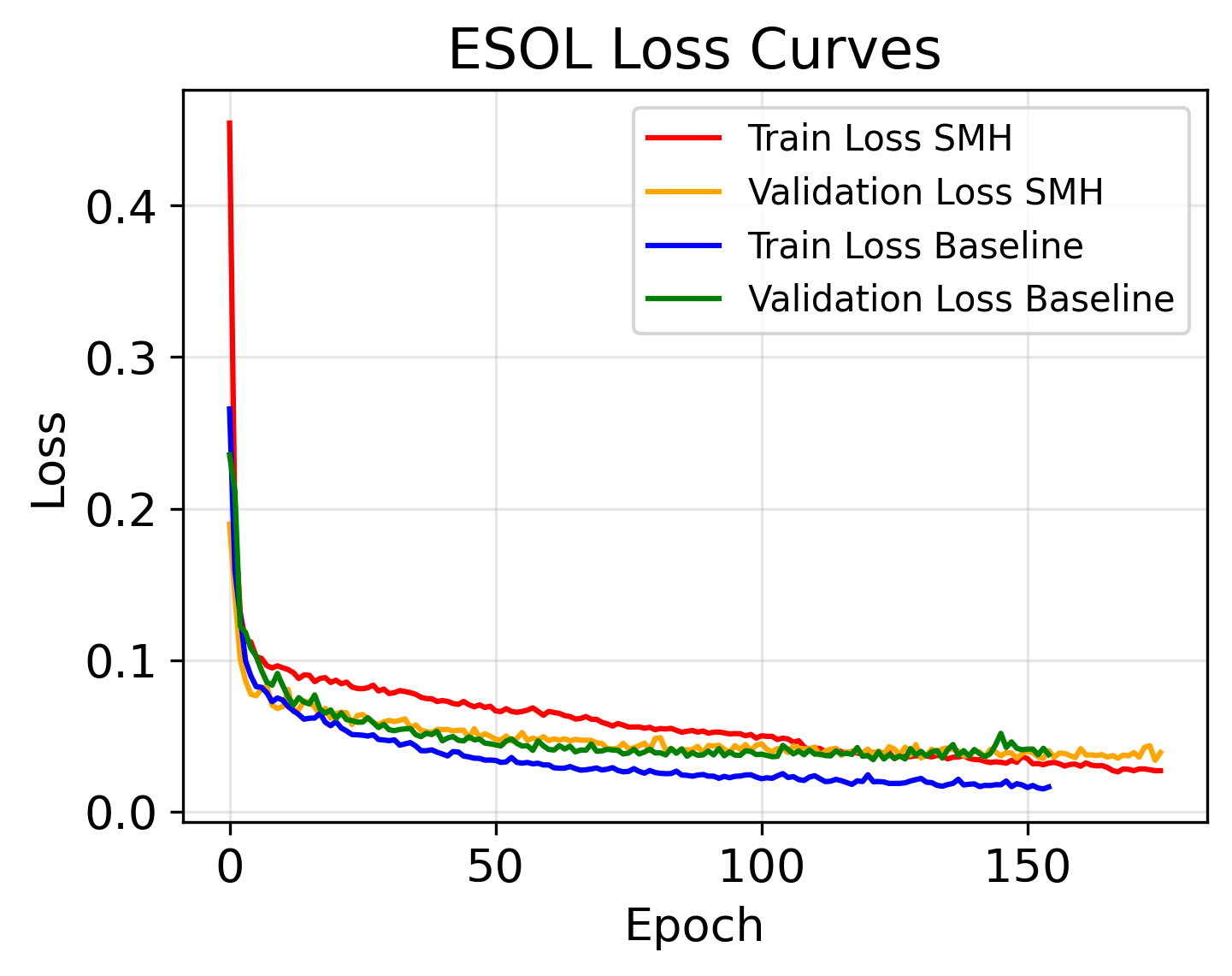}
\end{subfigure}
\begin{subfigure}{0.9\columnwidth}
    \includegraphics[width=1.0\linewidth]{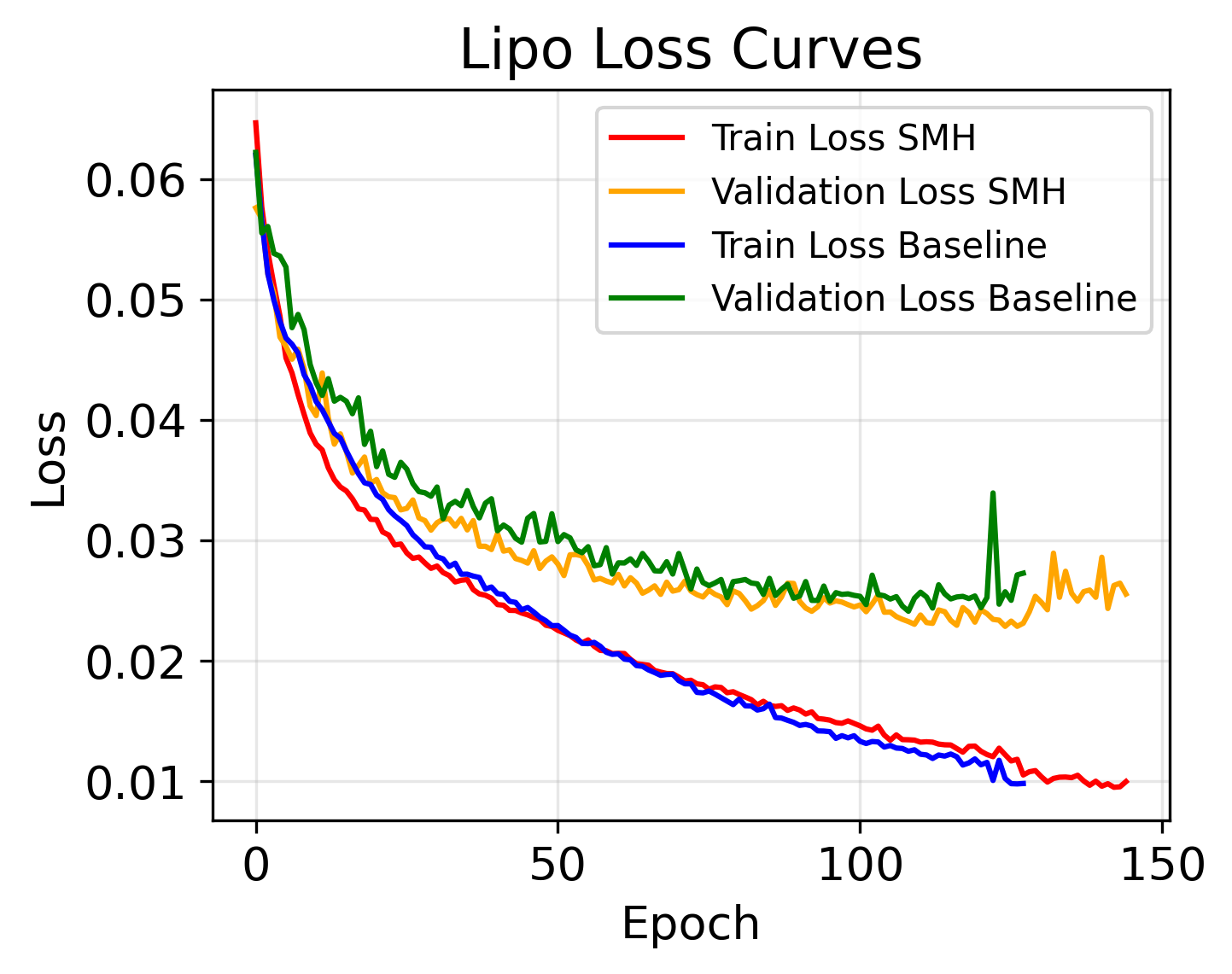}
\end{subfigure}
\caption{Training and validation performance of property value prediction for each dataset for original and augmented training sets.}
\label{fig:overral_perfom}
\end{figure}

The training and validation losses are presented in Figure~\ref{fig:overral_perfom}. 

%
%

\end{document}